\documentclass[runningheads]{llncs}
\usepackage[T1]{fontenc}
\usepackage{graphicx}
\usepackage{amsmath}
\begin{document}
\title{Lesion Segmentation in Moderate to Severe Traumatic Brain Injury: An nnU-Net Based Approach with Adaptive Normalization in the AIMS-TBI 2025 Challenge}
%
%
\author{Inhwa Son \inst{1} \and
Gaeun Lee\inst{1} \and
Sohyeon Sim\inst{1} \and
Kwang-Hyun Uhm\inst{1,2} 
} 
\authorrunning{F. Author et al.}
%
\institute{Gachon University, Republic of Korea \and
MEDAI, Republic of Korea \\
\email{\{inhwa1127, tong0430aa, thgus0101, khuhm\}@gachon.ac.kr}
}

\maketitle              
\begin{abstract}
The segmentation of lesions in Moderate to Severe Traumatic Brain Injury (msTBI) from T1-weighted MRI presents a significant clinical challenge due to the profound heterogeneity of lesion characteristics in terms of size, shape, and location. To address this, the AIMS-TBI 2025 Challenge was organized to promote the development of robust and accurate segmentation algorithms. In this paper, we present our deep learning-based solution. Our methodology employs the nnU-Net framework with an adaptive intensity normalization strategy confined to the brain parenchyma, effectively reducing inter-subject variability and mitigating artifacts from non-brain structures. 
Upon final evaluation on the held-out test set, our method demonstrated highly competitive performance on the official leaderboard, achieving an Overall Dice Coefficient of 0.6305. The model obtained a Dice score of 0.4805 for lesion segmentation and 0.9324 for non-lesion tissue. While the lesion Dice reflects the difficulty of detecting highly heterogeneous lesions, the high non-lesion Dice primarily indicates the model’s strong ability to correctly identify non-lesion voxels, demonstrating good specificity in differentiating lesion from non-lesion regions.
These results demonstrate that incorporating anatomically constrained normalization within the nnU-Net pipeline is a powerful and effective strategy for tackling the complexities of msTBI lesion segmentation.

\end{abstract}

\section{Introduction}
Traumatic brain injury (TBI) is a leading cause of mortality and long-term disability worldwide, arising from external mechanical forces such as traffic accidents, falls, or sports-related impacts. In moderate to severe TBI (msTBI), the rapid acceleration and deceleration of the brain within the skull induces both primary injuries (e.g., hematomas, hemorrhages, contusions) and secondary injuries (e.g., gliosis, encephalomalacia), which may necessitate urgent surgical intervention. These injuries produce diverse structural deformations in the brain, and each patient typically presents with a unique combination of lesion patterns. Such heterogeneity—spanning lesion size, number, laterality, and tissue involvement across gray matter, white matter, and cerebrospinal fluid—is widely recognized as a hallmark of msTBI.

This extreme heterogeneity introduces substantial challenges for neuroimaging analysis. Lesions in msTBI differ fundamentally from other pathologies such as stroke or tumors, as they may be focal or diffuse and frequently extend across multiple tissue types or bilateral regions. These complex patterns complicate image registration, normalization, and downstream parcellation, often leading to both local and global errors in brain analysis. While lesion compensation methods have been proposed—such as brain extraction or lesion inpainting—most approaches require labor-intensive manual segmentation, which is impractical in large-scale studies. Automated lesion segmentation tools developed for other etiologies have also shown limited performance in TBI, reflecting the need for specialized solutions.

In the absence of accurate automated tools, researchers have resorted to strategies such as ignoring lesions, excluding patients with large injuries, or relying on manual segmentation. These workarounds either reduce reliability, restrict generalizability, or limit statistical power for subgroup analyses. As a result, progress in understanding how factors such as lesion type, severity, or comorbid conditions influence patient outcomes has been constrained. To address these barriers, robust and scalable algorithms for msTBI lesion segmentation are urgently needed.
In recent years, deep learning methods have been increasingly applied to medical image analysis and have shown rapid progress, offering state-of-the-art performance across a wide range of tasks~\cite{npj_uhm,2022_uhm,UHM2024108746,kits23_uhm,TMI_uhm}.
This makes deep learning a particularly promising approach for addressing the unique challenges of lesion segmentation in msTBI.

The Automated Imaging for Moderate-to-Severe TBI (AIMS-TBI) Challenge was established to accelerate progress in this area. Leveraging multi-site data from the ENIGMA Consortium, the challenge focuses specifically on T1-weighted MRI—the most widely available modality across clinical cohorts—thereby enhancing applicability in large-scale, multi-institutional research. The inaugural AIMS-TBI Challenge, held at MICCAI 2024, demonstrated the feasibility of this task, with top-performing methods achieving a Dice score of 0.61. However, these results also underscored significant room for improvement, motivating subsequent editions with larger datasets and more diverse cohorts.

Building upon this foundation, the 2025 AIMS-TBI Challenge provides an expanded dataset and continues to benchmark lesion segmentation algorithms under realistic, multi-cohort conditions. Accurate automated segmentation of msTBI lesions will not only streamline neuroimaging workflows but also enable advanced analyses such as parcellation, functional and structural connectivity, and outcome prediction, ultimately contributing to improved clinical understanding and patient care. In this work, we present our solution to the AIMS-TBI 2025 Challenge, which leverages the nnU-Net framework enhanced with an adaptive intensity normalization strategy confined to the brain parenchyma. By incorporating anatomically constrained normalization into the preprocessing pipeline, our method reduces inter-subject variability and mitigates artifacts from non-brain structures, leading to more robust lesion segmentation performance on heterogeneous multi-site MRI data.

\section{Method}

\subsection{Dataset}
We utilized the dataset provided by the AIMS-TBI 2025 Challenge~\cite{dennis_2025_15084120}, which consists of multi-site 
T1-weighted (T1w) MRI scans of patients with moderate to severe traumatic brain injury (msTBI). 
The data were aggregated from 13 international sites participating in the ENIGMA Pediatric and Adult msTBI 
working groups, covering subjects aged 5–85 years (64\% male), with enrichment for adolescent cases to reflect 
epidemiological trends. Imaging was performed on 1.5T and 3T scanners from multiple manufacturers 
(GE, Siemens, Philips). Most scans had isotropic 1~mm$^3$ voxel resolution, though acquisition parameters varied 
within standard ranges.

A total of 875 MRI scans were included, split into 500 training (57\%), 100 validation (11\%), and 275 test (32\%) 
cases. A case was defined as one T1w MRI scan from a particular patient. While the majority of patients 
contributed a single scan, some contributed longitudinal scans (up to four time points); in these cases, all 
longitudinal scans were assigned to the same split (training, validation, or test) to prevent data leakage. 
The test set was designed to reflect real-world heterogeneity in age, sex, scanner, and lesion distribution.

Lesions encompassed a wide spectrum of injury-related pathologies, including contusions, hematomas, 
hemorrhages, encephalomalacia, gliosis, white matter lesions, and surgical drainage tracts. Reference lesion 
annotations were generated through a four-step process: (i) initial automated segmentation using a U-Net 
pretrained on the ATLAS v2.0 dataset, (ii) manual review and edits by a trained rater, (iii) secondary review 
by another rater, and (iv) final approval by an expert annotator. In total, seven primary and five expert annotators 
contributed to the labeling, following a standardized protocol. Annotation was performed in ITK-SNAP, and all 
raters underwent training with feedback to ensure consistency. Each image was reviewed by at least three 
annotators sequentially, minimizing missed lesions and boundary errors.

No lesions were excluded based on size; even very small lesions (e.g., $<10$ voxels) were retained during both 
training and evaluation. To protect privacy, all MRI scans were defaced using \texttt{pydeface}, and identifying 
metadata were removed. Only age (rounded to the nearest year) and time since injury (in weeks) were provided as demographic information.

\begin{figure*}[t]
    \centering
    \includegraphics[width=\textwidth]{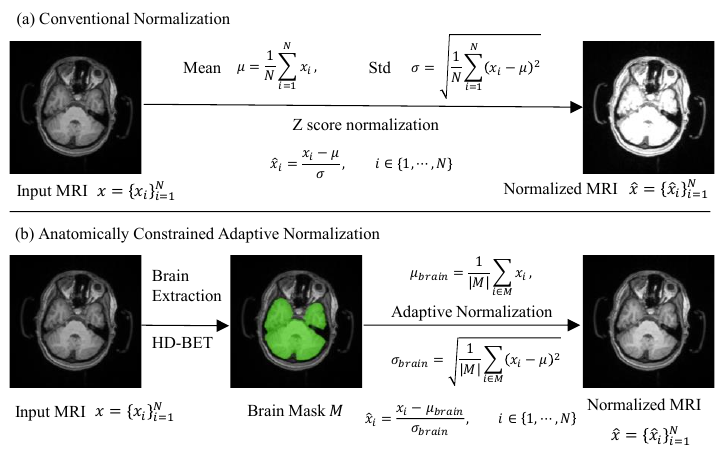}
    \caption{Comparison of normalization strategies within the nnU-Net framework. 
    (a) Conventional normalization computes mean and standard deviation over all voxels, 
    leading to potential variability from non-brain regions. 
    (b) The proposed anatomically constrained adaptive normalization restricts computation 
    to brain parenchyma voxels using a brain mask, thereby reducing inter-subject variance 
    and improving robustness.}
    \label{fig:framework}
\end{figure*}

\subsection{Preprocessing}
A key component of our method is an \textit{adaptive intensity normalization strategy} tailored to the heterogeneity of msTBI. In conventional preprocessing pipelines, normalization statistics (mean and standard deviation) are calculated over the entire image volume, including skull and background regions. This often results in substantial inter-subject variability, as skull brightness and non-brain tissues introduce inconsistencies across scans. Such variability can obscure lesion-specific signal characteristics and reduce the effectiveness of subsequent learning.  
As illustrated in Figure~\ref{fig:framework}, our approach differs from conventional normalization by restricting the computation of normalization statistics to the brain parenchyma.  

To mitigate this issue, we extracted the \textit{brain parenchyma mask} from each MRI scan. Normalization parameters were then computed exclusively within this intracranial region, and voxel intensities were normalized using these brain-specific statistics. Formally, given an input MRI $x = \{x_i\}_{i=1}^N$ with $N$ voxels and a binary brain mask $M$, we defined the masked mean and variance as  

\begin{equation}
\mu_{\text{brain}} = \frac{1}{|M|}\sum_{i \in M} x_i, \quad 
\sigma_{\text{brain}} = \sqrt{\frac{1}{|M|}\sum_{i \in M}(x_i - \mu_{\text{brain}})^2},
\end{equation}
where $\mu_{\text{brain}}$ and $\sigma_{\text{brain}}$ denote the mean and standard deviation 
computed within the brain mask $M$, and $|M|$ is the number of voxels inside the mask.
Normalized intensities were then obtained as  

\begin{equation}
\hat{x}_i = \frac{x_i - \mu_{\text{brain}}}{\sigma_{\text{brain}}}, \quad i = 1,\dots,N,
\end{equation}

By confining normalization to brain tissue, this strategy reduces variance across subjects caused by non-brain structures, leading to more stable intensity distributions and allowing the segmentation model to better focus on pathological regions. All MRI volumes were further resampled to isotropic $1~\text{mm}^3$ resolution and cropped or padded to standardized sizes as specified by the nnU-Net framework. 

We applied extensive data augmentation, including random rotations, scaling, flipping, intensity perturbations, elastic deformations, gamma corrections, and Gaussian noise, 
to enhance robustness to site-specific variability and improve generalization across heterogeneous cohorts.

\subsection{Model Architecture}
We adopted the \textit{Residual Encoder Large (resencL)} variant of the nnU-Net~\cite{nnunet} framework. This configuration employs residual blocks in the encoder path to improve feature representation and gradient flow, while maintaining the standard U-shaped architecture with skip connections and deep supervision. The large variant increases model capacity, making it suitable for capturing the heterogeneous and complex lesion patterns characteristic of msTBI.

\subsection{Training and Inference}
The network was trained end-to-end using a hybrid loss function combining \textit{Dice loss} and \textit{cross-entropy loss}, balancing overlap-based and voxel-wise optimization objectives. Optimization employed stochastic gradient descent with momentum, an initial learning rate of 0.01, and a polynomial decay schedule. Training was performed for 1000 epochs with extensive online data augmentation. 
In inference, \textit{test-time augmentation (TTA)} was applied by mirroring images along different spatial axes, and predictions were averaged across augmentations.

\section{Results}

\subsection{Evaluation Metrics}
The final evaluation was conducted on the hidden test set of the AIMS-TBI 2025 Challenge using four official metrics: 
(1) the mean position across all criteria, 
(2) the Dice coefficient for lesion segmentation, 
(3) the Dice coefficient for non-lesion tissue, 
and (4) the overall Dice coefficient, computed as the average of lesion and non-lesion Dice scores. 
This multi-metric evaluation ensured that algorithms were not only accurate in lesion segmentation but also reliable in preserving non-lesion regions.

\begin{table}[t]
\centering
\caption{Performance comparison across variants during preliminary development and final test leaderboard. Results highlight the effect of extensive data augmentation (DA5) and adaptive normalization (Adaptive Norm.).}
\label{tab:leaderboard_results}
\begin{tabular}{lcccc}
\hline
\textbf{Method} & \textbf{Balanced Acc.} & \textbf{Dice (les.)} & \textbf{Dice (no les.)} & \textbf{Overall} \\
\hline
\multicolumn{5}{c}{\textit{Preliminary Development Results (leaderboard)}} \\
\hline
Base (ResEnc L) & 0.8713 & 0.5514 & 0.8936 & 0.7123 \\
DA5 (extensive augmentation) & 0.8725 & 0.5428 & \textbf{0.9149} & 0.7177 \\
Adaptive Normalization & 0.8537 & 0.5450 & \textbf{0.9149} & \textbf{0.7189} \\
\textbf{DA5 + Adaptive Norm.} & \textbf{0.8808} & \textbf{0.5550} & 0.8936 & 0.7141 \\
\hline
\multicolumn{5}{c}{\textit{Final Test Results (leaderboard)}} \\
\hline
DA5 (extensive augmentation) & 0.8587 & \textbf{0.4820} & 0.9054 & 0.6225 \\
\textbf{DA5 + Adaptive Norm.} & \textbf{0.8622} & 0.4805 & \textbf{0.9324} & \textbf{0.6305} \\
\hline
\end{tabular}
\end{table}

\subsection{Effect of Adaptive Normalization and Data Augmentation}
Table~\ref{tab:leaderboard_results} summarizes the performance across different variants during both the preliminary development phase (Phase 1) and the final test leaderboard. The baseline model (“Base”) corresponds to ResEnc L without additional normalization or augmentation. Comparisons with “DA5” (extensive augmentation) and “Adaptive Norm.” (adaptive normalization) allow us to directly assess the contribution of each component.

During Phase 1, both DA5 and Adaptive Norm. independently improved robustness, particularly in terms of no-lesion Dice. Their combination (“DA5 + Adaptive Norm.”) achieved the highest overall score (0.8808), highlighting complementary benefits. These outcomes demonstrate that the proposed Adaptive Norm. provides measurable gains over no-normalization or augmentation-only strategies.

Final test results further confirm this trend: “DA5 + Adaptive Norm.” achieved the best overall Dice (0.6305), outperforming augmentation-only (0.6225). This consistency across development and test phases indicates that Adaptive Norm. enhances generalization, especially when combined with strong data augmentation.

\subsection{Leaderboard Results}
Table~\ref{tab:leaderboard} summarizes the performance of the top-ranking methods on the final test leaderboard. 
Our approach achieved a mean position of 5.5, with a lesion Dice of 0.4805, a non-lesion Dice of 0.9324, and an overall Dice of 0.6305. 
These results demonstrate that our adaptive normalization strategy, combined with the residual encoder large variant of nnU-Net, yields a balanced trade-off between lesion segmentation and preservation of non-lesion tissue. 

\begin{table}[ht]
\centering
\caption{Final test leaderboard results of the AIMS-TBI 2025 Challenge. Five metrics were reported: mean position, score, Dice for lesions, Dice for non-lesion tissue, and overall Dice.}
\label{tab:leaderboard}
\begin{tabular}{lccccc}
\hline
\textbf{Team} & \textbf{Pos.} & \textbf{Balanced Acc.} & \textbf{Dice (Les.)} & \textbf{Dice (No Les.)} & \textbf{Overall} \\
\hline
iMedIA\_2025     & 3.5  & \textbf{0.8924} & 0.4904 & \textbf{0.9324} & \textbf{0.6371} \\
AIMHI-MEDAI (ours)        & 5.5  & \underline{0.8622} & 0.4805 & \textbf{0.9324} & 0.6305 \\
NIC-VICOROB           & 6.0  & 0.8554 & 0.4941 & \underline{0.9189} & \underline{0.6351} \\
SpaceCY               & 8.3  & 0.8383 & \textbf{0.5149} & 0.8243 & 0.6176 \\ 
jianghaotian0001      & 8.3  & 0.8585 & \underline{0.5013} & 0.8243 & 0.6085 \\
\hline
\end{tabular}
\end{table}

\subsection{Performance Analysis}
Compared to other submissions, our method maintained highly competitive performance on both lesion and non-lesion Dice coefficients. 
Although the lesion Dice (0.4805) was modest relative to some teams, the strong non-lesion Dice (0.9324) contributed to a well-balanced overall Dice of 0.6305, placing our method among the top systems. 
These results validate the effectiveness of our anatomically constrained normalization strategy in reducing inter-subject intensity variance and improving robustness across heterogeneous multi-site MRI data.

Interestingly, while some methods achieved higher lesion Dice, this often came at the cost of reduced non-lesion accuracy, leading to more false positives or over-segmentation. 
In contrast, our approach prioritized robust representation of healthy tissue while maintaining reasonable lesion segmentation accuracy, yielding a well-balanced overall performance. 
This trade-off suggests that future improvements may come from enhancing sensitivity to small or diffuse lesions without sacrificing non-lesion fidelity.  

\section{Conclusion}
In this study, we presented a deep learning-based solution to the AIMS-TBI 2025 Challenge for lesion segmentation in moderate to severe TBI. 
Our method built upon the nnU-Net framework and introduced an adaptive normalization strategy restricted to the brain parenchyma, reducing inter-subject variance and improving robustness. 
By employing the residual encoder large (resencL) variant, and extensive data augmentation, our approach achieved strong performance on the official test leaderboard, with a balanced trade-off between lesion and non-lesion segmentation.  

These findings highlight the importance of anatomically informed preprocessing for robust lesion segmentation in heterogeneous clinical populations. 
In future work, we aim to further enhance lesion sensitivity, particularly for small and diffuse injuries, by incorporating multi-scale feature representations and integrating complementary MRI modalities. 
While our current analysis did not specifically examine performance variations across lesion size or anatomical location, we recognize this as an important direction for future investigation. 
Ultimately, accurate and automated lesion segmentation will facilitate advanced neuroimaging analyses and contribute to improved clinical understanding and prognostic modeling in msTBI.  

%
%
%
\bibliographystyle{splncs04}
\bibliography{refs}

@article{nnunet,
  title={nnU-Net: a self-configuring method for deep learning-based biomedical image segmentation},
  author={F. Isensee, P.F. Jaeger, S.A.A. Kohl et al.},
  journal={Nature Methods},
  volume={18},
  pages={203–-211},
  year={2021},
}

@misc{dennis_2025_15084120,
  author       = {Dennis, Emily and
                  Tustison, Nick and
                  Deutscher, Evelyn and
                  Wilde, Elisabeth and
                  Pease, Matthew and
                  Bakas, Spyridon},
  title        = {AIMS-TBI - Automated Identification of Moderate-
                   Severe Traumatic Brain Injury Lesions
                  },
  month        = mar,
  year         = 2025,
  publisher    = {Zenodo},
  doi          = {10.5281/zenodo.15084120},
  url          = {https://doi.org/10.5281/zenodo.15084120},
}

@article{npj_uhm,
author = {Kwang-Hyun Uhm and others},
title = {Deep learning for end-to-end kidney cancer diagnosis on multi-phase abdominal computed tomography},
journal = {npj Precis. Onc.},
volume = {5},
number = {54},
pages = {},
year = {2021},
month = {Jun.}
}

@ARTICLE{2022_uhm,
  author={Uhm, Kwang-Hyun and Jung, Seung-Won and Choi, Moon Hyung and Hong, Sung-Hoo and Ko, Sung-Jea},
  journal={IEEE Journal of Biomedical and Health Informatics}, 
  title={A Unified Multi-Phase CT Synthesis and Classification Framework for Kidney Cancer Diagnosis With Incomplete Data}, 
  year={2022},
  volume={26},
  number={12},
  pages={6093-6104}}

@article{UHM2024108746,
title = {Lesion-aware cross-phase attention network for renal tumor subtype classification on multi-phase CT scans},
journal = {Computers in Biology and Medicine},
volume = {178},
pages = {108746},
year = {2024},
author = {Kwang-Hyun Uhm and Seung-Won Jung and Sung-Hoo Hong and Sung-Jea Ko},}

@InProceedings{kits23_uhm,
author="Uhm, Kwang-Hyun
and Cho, Hyunjun
and Xu, Zhixin
and Lim, Seohoon
and Jung, Seung-Won
and Hong, Sung-Hoo
and Ko, Sung-Jea",
title="Exploring 3D U-Net Training Configurations and Post-processing Strategies for the MICCAI 2023 Kidney and Tumor Segmentation Challenge",
booktitle="Kidney and Kidney Tumor Segmentation",
year="2024",
publisher="Springer Nature Switzerland",
pages="8--13",
}

@ARTICLE{TMI_uhm,
  author={Uhm, Kwang-Hyun and Cho, Hyunjun and Hong, Sung-Hoo and Jung, Seung-Won},
  journal={IEEE Transactions on Medical Imaging}, 
  title={An Anisotropic Cross-View Texture Transfer with Multi-Reference Non-Local Attention for CT Slice Interpolation}, 
  year={2025},
  volume={},
  number={},
  pages={1-1},
  doi={10.1109/TMI.2025.3596957}}
\end{document}